\documentclass[conference]{IEEEtran}
\IEEEoverridecommandlockouts
\usepackage{flushend,cite}
\usepackage{amsmath,amssymb,amsfonts}
\usepackage{algorithmic}
\usepackage{graphicx}
\usepackage{textcomp}
\usepackage{xcolor}
\usepackage{booktabs}

\usepackage{subfigure}
\usepackage[hyphens]{url}
\usepackage{hyperref}
\usepackage{booktabs}

\usepackage{placeins}
\usepackage{float}
\restylefloat{figure}

\usepackage{sidecap}

\usepackage[hyphens]{url}
\usepackage{hyperref}
\usepackage{booktabs}
\usepackage{tikz}
\usetikzlibrary{positioning}
\usetikzlibrary{shapes.geometric}

\usepackage{balance}

\def\BibTeX{{\rm B\kern-.05em{\sc i\kern-.025em b}\kern-.08em
    T\kern-.1667em\lower.7ex\hbox{E}\kern-.125emX}}
\begin{document}

\title{Benchmarking of a new data splitting method\\ on volcanic eruption data
}
\makeatletter
\newcommand{\newlineauthors}{%
  \end{@IEEEauthorhalign}\hfill\mbox{}\par
  \mbox{}\hfill\begin{@IEEEauthorhalign}
}
\makeatother

\author{\IEEEauthorblockN{1\textsuperscript{st} Simona Reale}
\IEEEauthorblockA{\textit{Engineering Department} \\
\textit{University of Sannio}\\
Benevento, Italy \\
s.reale$@$studenti.unisannio.it}
\and

\IEEEauthorblockN{2\textsuperscript{nd} Pietro Di Stasio}
\IEEEauthorblockA{\textit{Engineering Department} \\
\textit{University of Sannio}\\
Benevento, Italy \\
p.distasio$@$studenti.unisannio.it}
\and

\IEEEauthorblockN{3\textsuperscript{rd} Francesco Mauro}
\IEEEauthorblockA{\textit{Engineering Department} \\
\textit{University of Sannio}\\
Benevento, Italy  \\
f.mauro$@$studenti.unisannio.it}
\and

\newlineauthors
\IEEEauthorblockN{4\textsuperscript{th} Alessandro Sebastianelli}
\IEEEauthorblockA{\textit{$\Phi$-lab} \\
\textit{European Space Agency}\\
Frascati, Italy \\ Alessandro.Sebastianelli@esa.int
}
\and

\IEEEauthorblockN{5\textsuperscript{th} Paolo Gamba}
\IEEEauthorblockA{\textit{Department of Electrical,
 Computer and Biomedical Engineering} \\
\textit{University of Pavia}\\
Pavia, Italy \\
paolo.gamba@universitadipavia.it}

\and

\IEEEauthorblockN{6\textsuperscript{th} Silvia Liberata Ullo}
\IEEEauthorblockA{\textit{Engineering Department} \\
\textit{University of Sannio}\\
Benevento, Italy \\
ullo@unisannio.it}
}

\maketitle

\begin{abstract}
In this paper, a novel method for data splitting is presented: an iterative procedure divides the input dataset of volcanic eruption, chosen as the proposed use case,  into two parts using a dissimilarity index calculated on the cumulative histograms of these two parts.  The  Cumulative Histogram Dissimilarity (CHD) index is introduced as part of the design. 
Based on the obtained results the proposed model in this case, compared to both Random splitting and K-means implemented over different configurations, achieves the best performance, with a slightly higher number of epochs. However, this demonstrates that the model can learn more deeply from the input dataset, which is attributable to the quality of the splitting. In fact, each model was trained with early stopping, suitable in case of overfitting, and the higher number of epochs in the proposed method demonstrates that early stopping did not detect overfitting, and consequently, the learning was optimal.
\end{abstract}

\begin{IEEEkeywords}
data splitting (DS), machine learning (ML), Cumulative Histogram Dissimilarity (CHD)
\end{IEEEkeywords}

\section{Introduction}
Data splitting is crucial in Machine Learning (ML) algorithms to prevent overfitting and underfitting. 
Proper data splitting ensures that the model is trained on a representative subset of the data and validated on another, leading to better generalization and more reliable predictions \cite{data_splitting_process}. 


Several data-splitting techniques are commonly used in ML
\cite{xu2018splitting, splitting_data, reitermanova2010data,  joseph2022split, birba2020comparative, Roshan}. One of the most straightforward methods is the train-test split, where the dataset is divided into two subsets: a training set and a testing set. Typically, 70-80\% of the data is used for training, while the remaining 20-30\% is reserved for testing \cite{splitting_data}. Although simple, this approach may not always provide a comprehensive evaluation of the model’s performance.

Another method is the train-validation-test split, which divides the dataset into three subsets: training, validation, and testing. The training set is used to train the model, the validation set helps to tune the hyperparameters and validate performance during the training, and the testing set evaluates the final model \cite{splitting_data}. This method is beneficial for fine-tuning the model but requires a larger dataset.

K-fold cross-validation is another technique where the dataset is divided into (k) equally sized folds. The model is trained and evaluated (k) times, each time using (k-1) folds for training and one fold for validation \cite{splitting_data}, \cite{SAFONOVA2023103569}. This technique provides a more robust estimate of model performance by reducing variance.

Stratified sampling  as other data splitting method ensures that the distribution of key features is preserved in both training and testing sets. This method is particularly useful for imbalanced datasets, where certain classes may have fewer samples \cite{splitting_data, joseph2022split}.


In addition to the classical methods, several alternative techniques have been proposed. One such technique is nested cross-validation, which involves an outer loop of cross-validation for model evaluation and an inner loop for hyperparameter tuning. This method provides an unbiased estimate of model performance and is particularly useful for small datasets \cite{data_splitting_process}. With the nested cross-validation, among alternative techniques,  bootstrap sampling can be mentioned. It involves repeatedly sampling with replacement from the dataset to create multiple training sets. Each sample is used to train the model, and the performance is averaged over all samples \cite{data_splitting_process}. This approach helps in estimating the variability of the model’s performance.
Another particular method is the Leave-One-Out Cross-Validation (LOOCV), where each data point is used once as a validation set while the remaining data points form the training set. Although computationally intensive, this method provides a thorough evaluation of model performance \cite{data_splitting_process}.
An interesting procedure is presented in \cite{wu2023split}, where the authors introduce a Split Learning (SL) framework, in which the data owner and the label owner each train distinct sections of the deep learning (DL) model, exchanging only the intermediate outputs. In this SL framework, the DL model is partitioned into two segments. During the training phase, the data owner, 
with access to the raw data, processes the first segment of the model and transmits the "smashed" data, which are the features derived from the raw data. The label owner, 
holding the label information, processes the second segment of the model and shares the gradients from the cut layer.  This learning collaborative paradigm shows to overcome previous limitations. 

The Splitting and Iterative Least Squares (SILS) training method is proposed in \cite{xi2021scalable}, to make the training process easy with large and high dimensional data. Because the least squares method can find pretty good weights during the first iteration, only a few succeeding iterations are needed to fine tune the Scalable Wide Neural Network (SWNN). Lastly, we can mention the Random Splitting method involving randomly partitioning the dataset into training, validation, and test sets. While simple to implement, it may not preserve the distributional properties of the data, leading to potential overfitting or underfitting \cite{rs12183054}.

Based on all the above, it is evident that benchmarking new data splitting methods on different Earth Observation (EO) tasks is essential to improve the predictive accuracy of ML models in geoscience. By comparing various techniques, researchers can identify the most effective methods for their specific datasets, leading to better-informed decisions and enhanced predictive capabilities.

\section{Method}
To introduce our method, let define $x_{(lat,lon)} \in \mathbf{R}^{(W,H,C)}$ a satellite image with a
width W, an height H and a number of channel equals to C acquired in a specific
region identified by the Latitude and Longitude coordinates,
respectively defined $lat$ and $lon$.
Let also define a collection of N samples $X \in \mathbf{R}^{(N,W,H,C)}$,
where N defines the dimensionality of the collection, here
defined as dataset.
\begin{equation} 
\centering
\forall x_{(lat,lon)} \in \mathbf{X}
\end{equation} 
The scope is to find a procedure for splitting the dataset
into three sub-datasets here defined as $X_{train}$, $X_{val}$ and $X_{test}$ respectively belonging to
$\mathbf{R}^{(M,W,H,C)}$, $\mathbf{R}^{(P,W,H,C)}$ $\mathbf{R}^{(Q,W,H,C)}$ where: 
\begin{equation}
\centering
N \ge M + P + Q
\label{Eq}
\end{equation}
where $N, M, P, Q \in Z^{+}$ and typically $M > P$ e $M > Q$.
It is important to note the equation \ref{Eq} presents the greater
than equal symbol because are rounded number calculated as
fraction of N as follow:
\begin{equation} M = \alpha N  \end{equation}
\begin{equation} P = \beta N  \end{equation}
\begin{equation} Q = \gamma N  \end{equation}
Typical values of $\alpha$, $\beta$ e $\gamma$ are respectively 0,8, 0,1 e 0,1.
In the specific case of our application, the volcanic eruption was considered and a related dataset selected for evaluation, as better specified in the next section. In particular, for time series data, such as volcanic eruption records, a time-based split is used. The dataset is usually split chronologically. This helps in evaluating the model’s performance on future unseen data \cite{splitting_data}.

The new method proposed in this paper is an iterative method dividing the input dataset into two parts using a dissimilarity index calculated on the cumulative histograms of these two parts. 
 The  Cumulative Histogram Dissimilarity (CHD) index is introduced and is a part of the method described ahead.  
 The method is applied to the input dataset to obtain the initial \textbf{train-validation split}; then the same procedure is again applied to the training dataset to obtain, along with the first split, the final \textbf{train-validation-test split}. This procedure is detailed below.

\begin{itemize}
\item \textbf{Train-validation split}
The proposed algorithm iteratively creates a random split of the input dataset to generate the train-validation split $X_{(i)train}$ and $X_{(i)val}$, where \(i\) represents the current iteration. Note that $\alpha$ and $\beta$ are fixed (e.g., $\alpha = 0,9$, $\beta = 0,1$). \(K\) images are randomly extracted from $X_{(i)train}$ and $X_{(i)val}$ to calculate the cumulative histograms $h_{(i)train}$ e $h_{(i)val}$. Note that \(K\) must satisfy the following rule: 
\begin{equation}
\centering
K \le min(M, P)
\end{equation}
Therefore, the dissimilarity index can be calculated as follows: 
\begin{equation}
     \centering
     d^{(i)} = \frac{\left|\frac{1}{K}\sum\frac{h^{(i)}_{train}}{\#
bins}-\frac{1}{K}\sum\frac{h^{(i)}_{val}}{\#
bins}\right|}{\frac{1}{K}\sum\frac{h^{(i)}_{train}}{\# bins}}
\end{equation}
Note that the division by $\#bins$ and by $\frac{1}{K}\sum\frac{h^{(i)}_{train}}{\# bins}$ is used to normalize the dissimilarity index within a reasonable range. By minimizing the dissimilarity vector $d \in R^{S}$, with \(S\) equal to the number of iterations, the train-val split with the lowest dissimilarity value can be extracted, or in other words, the most similar $X_{train}$ and $X_{val}$ can be obtained. A similar procedure is applied to the new 
$X_{train}$ to obtain the train-test split.

\item \textbf{Train-test split}
Similarly, the algorithm creates a random split of the input dataset, which in this case is the training dataset, to generate the train-test split $X_{(i)train}$ and $X_{(i)test}$.
Note that \(K\) must satisfy the following rule:  
\begin{equation}
\centering
K \le min(M^*, Q)
\end{equation} 
where $M^*$ is the new size of the training dataset after the first splitting procedure. 

Therefore, the dissimilarity index can be calculated as follows:
\begin{equation}
     \centering
     d^{(i)} = \frac{\left|\frac{1}{K}\sum\frac{h^{(i)}_{train}}{\#
bins}-\frac{1}{K}\sum\frac{h^{(i)}_{test}}{\#
bins}\right|}{\frac{1}{K}\sum\frac{h^{(i)}_{train}}{\# bins}}
\end{equation}
Similarly, the procedure is carried out as described above. By combining the train-val and the train-test split the final
train-val-test split is obtained. It is important to note that the decision to divide this procedure into two phases was made intentionally.\\
Indeed, in the case where $X_{val} = X_{test}$, it is sufficient to execute only the first part of the proposed algorithm.
\end{itemize}

\begin{figure}[!ht]
\centering
\includegraphics[scale=0.55]{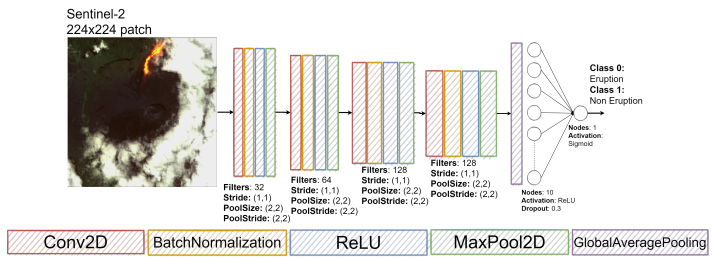}
\caption{Neural network architecture \cite{del2021board}}
\label{ReteArchitettura}
\end{figure}

\section{Dataset}
The accurate prediction of volcanic eruptions is a critical task in EO, necessitating robust ML models trained on historical eruption data. To benchmark the proposed data splitting technique, it was considered the task of vulcanic eruption clssification, on the dataset proposed by Del Rosso et al. \cite{del2021board}. This dataset was constructed by selecting some volcanic eruptions documented in the Volcanoes of the World (VOTW) catalog \cite{catalog}, which includes geolocation information on vulcanic eruptions.
An example of the information available in this catalog is shown in Table \ref{eruption_details}. 

\begin{table}[!ht]
\caption{Information in the Volcanoes of the World
(VOTW) catalog}
    \label{eruption_details}
\centering
\begin{tabular}{cccc}
\hline
\textbf{\begin{tabular}[c]{@{}c@{}}Eruption\\  Start Time\end{tabular}} & \textbf{\begin{tabular}[c]{@{}c@{}}Volcano\\  Name\end{tabular}} & \textbf{\begin{tabular}[c]{@{}c@{}}Latitude\\  (deg)\end{tabular}} & \textbf{\begin{tabular}[c]{@{}c@{}}Longitude\\  (deg)\end{tabular}} \\ \hline
26 June 2019                                                            & Ulawun                                                           & -5.050                                                             & 151.330                                                             \\
24 June 2019                                                            & Ubinas                                                           & -16.355                                                            & 151.330                                                             \\
22 June 2019                                                            & Raikoke                                                          & 48.292                                                             & 153.250                                                             \\
11 June 2019                                                            & \begin{tabular}[c]{@{}c@{}}Piton de \\ la Fournaise\end{tabular} & -21.244                                                            & 55.708                                                              \\
1 June 2019                                                             & Great Sitkin                                                     & 52.076                                                             & -176.130                                                            \\ \hline
\end{tabular}
    \label{tab:eruption_details}
\end{table}

For the dataset building, Landsat-7 and Sentinel-2 data were collected:
Landsat-7 data are considered for the period 1999-2015, while Sentinel-2 data for the period 2015-2019.


In Table \ref{tabellaLandsatSentinel} are hilighted  the differences between Sentinel-2 and Landsat-7 products in terms of spatial resolution and bandwidth.

\begin{table}[!ht]
\caption{Main characteristics of Landsat-7 and Sentinel-2}
    \label{tabellaLandsatSentinel}
\centering
\resizebox{\columnwidth}{!}{%
\begin{tabular}{|cccc|cccc|}
\hline
\multicolumn{4}{|c|}{\textbf{Landsat-7}} & \multicolumn{4}{c|}{\textbf{Sentinel-2}} \\ \hline
\multicolumn{1}{|c|}{\textbf{Band}} & \multicolumn{1}{c|}{\textbf{\begin{tabular}[c]{@{}c@{}}Wavelength \\ (nm)\end{tabular}}} & \multicolumn{1}{c|}{\textbf{\begin{tabular}[c]{@{}c@{}}Bandwidth \\ (nm)\end{tabular}}} & \textbf{\begin{tabular}[c]{@{}c@{}}Spatial resolution \\ (nm)\end{tabular}} & \multicolumn{1}{c|}{\textbf{Band}} & \multicolumn{1}{c|}{\textbf{\begin{tabular}[c]{@{}c@{}}Wavelength \\ (nm)\end{tabular}}} & \multicolumn{1}{c|}{\textbf{\begin{tabular}[c]{@{}c@{}}Bandwidth \\ (nm)\end{tabular}}} & \textbf{\begin{tabular}[c]{@{}c@{}}Spatial resolution \\ (nm)\end{tabular}} \\ \hline
\multicolumn{1}{|c|}{B1 (Blue)} & \multicolumn{1}{c|}{485} & \multicolumn{1}{c|}{70} & 30 & \multicolumn{1}{c|}{B2 (Blue)} & \multicolumn{1}{c|}{496.6(S2A)/492.1(S2B)} & \multicolumn{1}{c|}{66} & 10 \\ \hline
\multicolumn{1}{|c|}{B2 (Green)} & \multicolumn{1}{c|}{560} & \multicolumn{1}{c|}{80} & 30 & \multicolumn{1}{c|}{B3 (Green)} & \multicolumn{1}{c|}{560(S2A)/559(S2B)} & \multicolumn{1}{c|}{36} & 10 \\ \hline
\multicolumn{1}{|c|}{B3 (Red)} & \multicolumn{1}{c|}{660} & \multicolumn{1}{c|}{70} & 30 & \multicolumn{1}{c|}{B4 (Red)} & \multicolumn{1}{c|}{664.5(S2A)/665(S2B)} & \multicolumn{1}{c|}{31} & 10 \\ \hline
\multicolumn{1}{|c|}{B5 (SWIR1)} & \multicolumn{1}{c|}{1650} & \multicolumn{1}{c|}{200} & 30 & \multicolumn{1}{c|}{B11 (SWIR1)} & \multicolumn{1}{c|}{1613.7(S2A)/1610.4(S2B)} & \multicolumn{1}{c|}{91(S2A)/94(S2B)} & 20 \\ \hline
\multicolumn{1}{|c|}{B7 (SWIR2)} & \multicolumn{1}{c|}{2220} & \multicolumn{1}{c|}{260} & 30 & \multicolumn{1}{c|}{B12 (SWIR2)} & \multicolumn{1}{c|}{2202.4(S2A)/2185.7(S2B)} & \multicolumn{1}{c|}{175(S2A)/185(S2B)} & 20 \\ \hline
\end{tabular}%
}
\label{tab: landsat_s2}
\end{table}

The patches of the dataset cover a total area of $56.25$ $km^2$.
The original images, 
are resized to 512 x 512 pixels.
This resizing, is performed using the Python OpenCV library with the Bicubic Interpolation method.
This approach helps to reduce the spatial resolution differences between Sentinel-2 and Landsat-7 images while maintaining image quality.
To extract relevant information on volcanic eruptions, SWIR 
and RGB bands are considered.
Eruptions can be localized using a specific combination of RGB wavelength bands \cite{di2022early} captured by the satellite camera during the eruptive event.
Infrared bands are necessary to highlight high-temperature ground.
The infrared bands were then combined with the RGB bands to visually highlight the color of volcanic lava.
The combination of bands is given by \cite{Active_Volcanoes}:

\begin{subequations}
    \centering
    \begin{align}
        B4_n &= \alpha_1\cdot B4 + \max(0, SWIR2 - 0.1)\\
        B3_n &= \alpha_2\cdot B3 + \max(0, SWIR1 - 0.1)\\
        B2_n &= \alpha_3\cdot B2
    \end{align}
\end{subequations}

The multiplicative factor $\alpha_x$ is used to adjust the image scale and is set to $2,5$ \cite{del2021board}.
These bands are used because, during the acquisition phase, it is possible that the eruption has already subsided for some time, causing the lava to appear dark on the outside due to solidification, though still hot on the inside.
By using thermal infrared bands, the lava flow can be highlighted due to the presence of high heat.
Furthermore, for the dataset creation three images for each of the five selected volcanoes are downloaded via GEE, specifically pre-eruption, during-eruption and post-eruption images, with the most significant ones chosen to distinguish between the various periods.

By adopting these techniques, it is possible to create a quantitatively accurate dataset, as eruptions are easily distinguishable from non-eruptions during labeling.
Figure \ref{diff_immaginiRGBIR} shows the difference between a simple RGB image and one that highlights SWIR.

\begin{figure}[!ht]
    \centering
    \includegraphics[scale = 0.5]{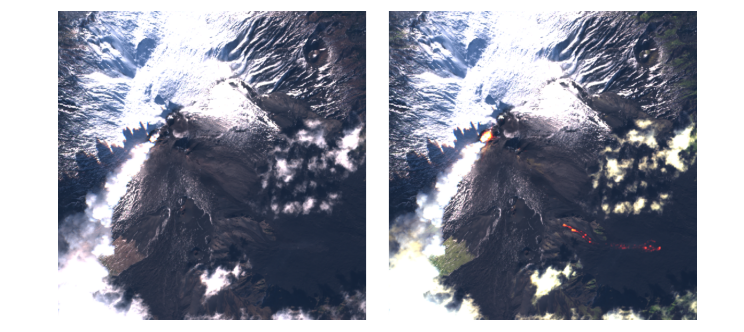}
    \caption{Real RGB color image (left) and IR-enhanced image (right) \cite{del2021board}}
    \label{diff_immaginiRGBIR}
\end{figure}

In defining the proposed dataset, it is necessary to consider two different classes: "eruption" and "non-eruption".
To ensure high variability and achieve better results, the images corresponding to non-eruptions are downloaded considering five subclasses: non-erupting volcanoes, cities, mountains, cloudy images, and completely random images.

\section{Results}


This section presents the results of the proposed method and comparisons with classical techniques, using the same dataset and CNN model but varying the splitting techniques. Several tables show the training time, the number of epochs, and the confusion matrix for each splitting method. The confusion matrix, or misclassification table, represents the accuracy of statistical classification, highlighting where the model makes errors and in which instances it performs better or worse. Each column refers to the results of the splitting algorithm under particular settings. The aim is to compare classical methods with the proposed one, 
by varying the settings, 
to demonstrate that our method yields better results with a significant margin. Table \ref{tabella1} shows the results of tests performed with the Random splitting (and two splitting factors of 0.1 and 0.15 are randomly chosen).

\begin{table}[!ht]
    \caption{Test with Random Splitting and splitting factor 0.1 and 0.15}
    \label{tabella1}
    \centering
    \begin{tabular}{|l|c|c|}
    \hline
    & Random splitting 0.15 & Random splitting 0.1\\
    \hline
    Time & $945,902 s$ & $1481,545 s$ \\
    \hline
    Epochs & $27/100$ & $42/100$ \\
     \hline
     True positive & $0,869$ & $0,923$\\
     \hline
     False positive & $0,131$ & $0,077$\\
     \hline
     False negative & $0,154$ & $0,225$\\
     \hline
     True negative & $0,846$ & $0,775$\\
     \hline
\end{tabular}
\end{table}

Table \ref{tabella2} shows the results of tests performed with K-means Cross Validation with 5 folds. Table \ref{tabella4} shows the results of tests performed with the Cumulative Histogram Dissimilarity (CHD) splitting in various configurations. Initially, tests were conducted with some CHD iterations equal to 50 and 100, both with a splitting factor of 0.2. Subsequently, iteration 100, which provided better results, was considered instead of 50, and further tests were conducted with splitting factors of 0.15 and 0.1.

\begin{table}[!ht]
    \caption{Test on K-means Cross Validation with 5 folds}
    \label{tabella2}
    \centering
    \resizebox{1\columnwidth}{!}{
\begin{tabular}{|l|c|c|c|c|c|}
    \hline
    K-means 5 & Fold 0 & Fold 1 & Fold 2 & Fold 3 & Fold 4\\
    \hline
    Time & $1949,784 s$ & $882,241 s$ & $1368,649 s$ & $1047,396 s$ & $913,847 s$ \\
    \hline
    Epochs & $51/100$ & $21/100$ & $37/100$ & $31/100$ & $24/100$\\
     \hline
     True positive & $0,868$ & $0,923$ & $0,916$ & $0,853$ & $0,909$\\
     \hline
     False positive & $0,132$ & $0,076$ & $0,084$ & $0,147$ & $0,090$\\
     \hline
     False negative & $0,234$ & $0,137$ & $0,167$ & $0,116$ & $0,347$\\
     \hline
     True negative & $0,765$ & $0,862$ & $0,833$ & $0,884$ & $0,653$\\
     \hline
\end{tabular}}
\end{table}

\begin{table}[!ht]
    \caption{Test with Cumulative Histogram Dissimilarity splitting with various configurations}
    \label{tabella4}
    \centering
    \resizebox{1\columnwidth}{!}{
\begin{tabular}{|l|c|c|c|c|}
     \hline
     & CHD iteration 50(split 0.2) & CHD iteration 100(split 0.2) & CHD best(100)(split 0.15) & CHD best(100)(split 0.1) \\
    \hline
    Time & $1732,034 s$ & $1708,945 s$ & $1810,033 s$ & $1923,356 s$\\
    \hline
    Epochs & $50/100$ & $50/100$ & $48/100$ & $50/100$\\
     \hline
     True positive & $0,885$ & $0,896$ & $0,897$ & $0,934$\\
     \hline
     False positive & $0,115$ & $0,104$ & $0,103$ & $0,066$\\
     \hline
     False negative & $0,160$ & $0,136$ & $0,131$ & $0,125$\\
     \hline
     True negative & $0,840$ & $0,864$ & $0,869$ & $0,875$\\
     \hline
\end{tabular}}
\end{table}

\section{Discussions and Conclusions}
Based on the results reported in the tables above, the proposed model, compared to both Random splitting and K-means carried out in different configurations, achieves the best performance practically always for the test with CHD, 100 iterations, splitting 0.1, even though it took more time for learning in terms of seconds and epochs. Although this last aspect might seem not completely positive, instead the reasoning behind that is related to the ability of the model to learn (with little more time and epochs)  more deeply from the input dataset, and this is attributable to the quality of the splitting. In fact, each model was trained with early stopping, a method that allows stopping the learning process in case of overfitting, which means that for the proposed method, early stopping did not detect overfitting, and consequently, the learning was optimal.

\bibliographystyle{IEEEtran}
\bibliography{ref}
\end{document}